\documentclass{article}

    \PassOptionsToPackage{numbers, compress}{natbib}

\usepackage[preprint]{neurips_2025}

\usepackage[utf8]{inputenc} 
\usepackage[T1]{fontenc}    
\usepackage{hyperref}       
\usepackage{url}            
\usepackage{booktabs}       
\usepackage{amsfonts}       
\usepackage{nicefrac}       
\usepackage{microtype}      
\usepackage{xcolor}         

\usepackage{amsmath}
\usepackage{graphicx}
\usepackage{algorithm}
\usepackage{algorithmic}
\usepackage{multirow}
\usepackage{bbding}
\usepackage{enumitem}
\usepackage{wrapfig}
\usepackage{authblk}
\usepackage{latexsym}

\title{CSDformer: A Conversion Method for Fully Spike-Driven Transformer}

\makeatletter
\def\thanks#1{\protected@xdef\@thanks{\@thanks
        \protect\footnotetext{#1}}}
\makeatother
\author[1]{Yuhao Zhang}
\author[1]{Chengjun Zhang}
\author[2]{Di Wu}
\author[2*]{Jie Yang \thanks{*Corresponding authors: Jie Yang and Mohamad Sawan}}
\author[1,2*]{Mohamad Sawan}
\affil[1]{Zhejiang Key Laboratory of 3D Micro/Nano Fabrication and Characterization, \authorcr Westlake Institute for Optoelectronics, Fuyang, Hangzhou, Zhejiang 311421, China}
\affil[2]{CenBRAIN Neurotech, School of Engineering, 
\authorcr Westlake University, Hangzhou, Zhejiang 310030, China.}

\begin{document}

\maketitle

\begin{abstract}
Spike-based transformer is a novel architecture aiming to enhance the performance of spiking neural networks while mitigating the energy overhead inherent to transformers. However, methods for generating these models suffer from critical limitations: excessive training costs introduced by direct training methods, or unavoidably hardware-unfriendly operations in existing conversion methods. In this paper, we propose CSDformer, a novel conversion method for fully spike-driven transformers. We tailor a conversion-oriented transformer-based architecture and propose a new function NReLU to replace softmax in self-attention. Subsequently, this model is quantized and trained, and converted into a fully spike-driven model with temporal decomposition technique. Also, we propose delayed Integrate-and-Fire neurons to reduce conversion errors and improve the performance of spiking models. We evaluate CSDformer on ImageNet, CIFAR-10 and CIFAR-100 datasets and achieve 76.36\% top-1 accuracy under 7 time-steps on ImageNet, demonstrating superiority over state-of-the-art models. Furthermore, CSDformer eliminates the need for training SNNs, thereby reducing training costs (reducing computational resource by 75\% and accelerating training speed by 2-3$\times$). To the best of our knowledge, this is the first fully spike-driven transformer-based model developed via conversion method, achieving high performance under ultra-low latency, while dramatically reducing both computational complexity and training overhead.
\end{abstract}

\section{Introduction}
\label{sec:introduction}
Spiking neural networks (SNNs), as the third generation of neural networks, have found widespread application across various tasks \cite{wu2018spiking,imam2020rapid,kim2020spiking,he2020comparing}. Unlike traditional artificial neural networks (ANNs), the event-driven nature makes SNNs more biologically plausible, and confers significant advantages in terms of low energy consumption \cite{roy2019towards,schuman2022opportunities}. Transformer, a novel network architecture, has achieved remarkable advancements in the field of deep learning \cite{vaswani2017attention,dosovitskiy2020image}. Consequently, there is a growing trend toward integrating the benefits of SNNs and transformers to create innovative models.

Inspired by ANNs, methods for generating SNNs can be categorized into two main approaches: direct training (DT) and ANN-to-SNN conversion. For DT methods, surrogate gradient is used to solve the non-differentiable of spikes \cite{neftci2019surrogate}. At the same time, with the development of deep learning, SNNs are also deeper and more effective \cite{fang2021deep,su2023deep}. Furthermore, the performance of spike-based transformers based on DT methods has surpassed traditional architectures \cite{zhou2022spikformer,zhou2023spikingformer,yao2024spike}, but they often require high costs to be trained. For example, the largest model in Spikingformer is trained about 2 weeks \cite{zhou2023spikingformer}.

Another approach to generate deep SNNs is ANN-to-SNN conversion. The pre-trained ANN is converted to an SNN by replacing activation functions with spiking neurons. In the preliminary stage, statistics-based methods significantly reduce training costs by avoiding retraining ANNs, but the resulting SNNs exhibit high latency and low performance \cite{kim2020spiking,rueckauer2017conversion,sengupta2019going}. In recent years, through the modification and retraining of ANNs, the converted SNNs have achieved ultra-low latency and competitive performance \cite{deng2021optimal,bu2023optimal}. Some conversion methods are also applied to transformers, but those converted spiking transformers still retain normalizations during inference, such as softmax and layer normalization \cite{mueller2021spiking,wang2023masked,you2024spikezip}. As a result, they cannot be deployed on neuromorphic chips \cite{merolla2014million,davies2018loihi} to fully leverage the high energy efficiency of SNNs.

In conclusion, current methods for spike-based transformers face serious challenges: direct training approaches suffered from expensive training overhead, whereas conversion-based paradigms retained hardware-unfriendly normalization operations. In this paper, we propose a novel conversion method for generating fully spike-driven transformers, especially for vision tasks, with only hardware-friendly operations. We train quantized transformers based on conversion-oriented architectures, and then convert them into fully spike-driven models. This method can thus save a lot of training costs without training SNNs. The main contributions of this paper are as follows:
\begin{itemize}[leftmargin=*]
    
    \item We propose a novel ANN-to-SNN conversion method to generate fully spike-driven transformers, named CSDformer. This method significantly reduces training costs, while ensuring the converted model operates through hardware-friendly, spike-driven computations.
    \item We propose NReLU to replace the softmax in self-attention, thereby eliminating normalization operations while maintaining hardware compatibility. Furthermore, we propose delayed Integrate-and-Fire neuron model, on the top of temporal decomposition technique, to reduce conversion errors and improve the performance of spike-driven models.
    \item Extensive experiments demonstrate that our CSDformers can achieve superior or comparable performance compared to state-of-the-art SNNs, with hardware-friendly operations and less training costs. We achieve 76.36\% top-1 accuracy under 7 time-steps on ImageNet, while reducing computational resource by 75\% and accelerating training speed by 2-3$\times$.
\end{itemize}

\section{Related Works}

\subsection{Vision Transformers}
Vision Transformer (ViT) represents an typical application in computer vision, specifically for image classification \cite{dosovitskiy2020image}. A classic ViT includes a patch splitting module (tokenizer), a transformer encoder, and a linear head. As transformers continue to evolve, ViT and its variants have achieved outstanding performance on a variety of vision benchmarks \cite{liu2021swin,rao2021dynamicvit,zhang2021vit}. Nevertheless, owing to the large model size and complicated computational demands, there is a critical need to develop a hardware-friendly and energy-efficient ViT \cite{mehta2021mobilevit,cai2023efficientvit}. Motivated by these models, we focus on a combination of SNN and ViT, and propose an energy-efficient spike-driven transformer for image classification.

\subsection{Spike-based Transformers}

Most existing architectures of SNNs inspired by convolutional neural networks (CNNs), such as VGG and ResNet, thus their performance is limited by CNNs \cite{fang2021deep,sengupta2019going}. The significant breakthrough brought by the attention-based transformer has led researchers to consider how to integrate this architecture into SNNs. Some researchers have proposed models based on direct training methods, including Spikformer \cite{zhou2022spikformer}, Spikingformer \cite{zhou2023spikingformer}, Spike-driven Transformer \cite{yao2024spike} and so on. However, due to the design limitations of GPUs, training these models requires high costs, including training time, computing and memory resources. Several other projects have focused on generating spike-based transformers through conversion methods. Spiking Transformer Networks was the first one to incorporate spiking neurons into the network \cite{mueller2021spiking}, while Masked Spiking Transformer (MST) introduced a random mask to reduce energy consumption without performance degradation \cite{wang2023masked}. SpikeZIP-TF converted a quantized transformer into spike-based model with Spike-Softmax and Spike-LayerNorm \cite{you2024spikezip}. Although these methods convert activation functions into spiking neurons, they still involve complicated operations, such as division and exponentiation in softmax, which make it difficult to achieve fully spike-driven computation and prevent deploy them on resource-limited neuromorphic chips. Spatio-Temporal Approximation proposed spatial and temporal approximations to convert a standard model \cite{jiang2024spatio}. However, this method heavily relies on effective approximators, and the whole framework is quite complicated. To sum up, it is urgent to develop a clear yet simple conversion framework for fully spike-driven transformers, which is also the goal of this work.

\section{Conversion Method}
Our framework starts with a tailored transformer, which inherently eliminates normalization operations during inference, thereby facilitating fully spike-driven calculations. Then, we leverage quantization-aware-training (QAT) technique to quantize this model, and execute a deterministic conversion for spike-driven model generation. In the following sections, we provide a systematic exposition of the architecture and core components of CSDformer.

\subsection{Tailored Transformer}
\label{sec:tailoredformer}
\subsubsection{Overall Architecture}
\label{sec:architecture}
\begin{figure}
    \centering
    \includegraphics[width=\textwidth]{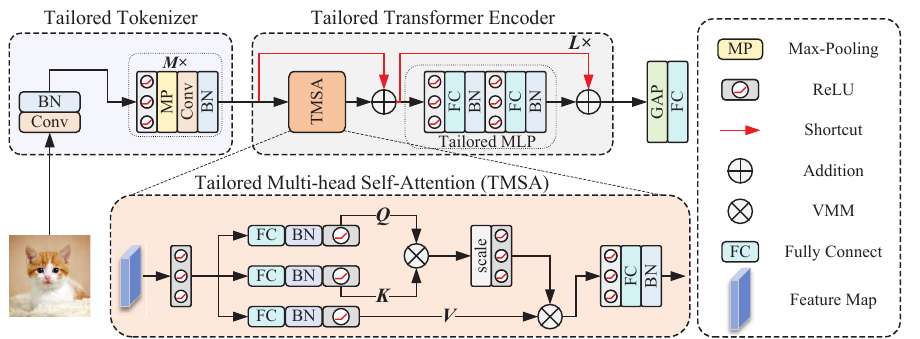}
    \caption{Architecture of Tailored Transformer. Based on the model established in \cite{zhou2023spikingformer}, we substantially redesign the standard ViT through three key innovations. First, we use a pure convolutional module for tokenization. Second, we systematically substitute all LayerNorm and activation functions with BatchNorm and ReLU. Also, BatchNorm is applied after and ReLU is applied before each linear transformation, such as convolutional and fully-connected layers. Finally, we propose a new function, named NReLU, to replace softmax in self-attention for avoiding complex operations, such as division and exponentiation. BN: Batch Normalization; VMM: variable matrix multiplication.}
    \label{fig:tailoredvit}
\end{figure}

First of all, we refer to the model in \cite{zhou2023spikingformer} to design a conversion-oriented architecture, as shown in Fig.~\ref{fig:tailoredvit}. Similar to standard ViT, the tailored model consists of a tailored tokenizer (TT), and several transformer blocks, each including a tailored multi-head self-attention (TMSA) and a tailored multilayer perceptron (TMLP). A global average-pooling (GAP) and a fully-connected classification head (CH) are utilized to output the prediction $Y$. We present a pure convolutional tokenizer to facilitate the conversion of spike-driven model and implementation on neuromorphic hardware. The tailored transformer can be calculated as follows:
\begin{align}
    X_0&=\mathrm{TT}(I), \quad I\in\mathbb{R}^{C \times H \times W}, \quad X_0\in\mathbb{R}^{N \times D} \\
    X^\prime_l&=\mathrm{TMSA}(X_{l-1})+X_{l-1}, \quad l=1 \ldots L \\
    X_l&=\mathrm{TMLP}(X^\prime_l)+X^\prime_l, \quad \quad ~~~~ l=1 \ldots L \\
    Y&=\mathrm{CH}(\mathrm{GAP}(X_L))
\end{align}

\subsubsection{Activation and Normalization}
\label{sec:relu_bn}

Almost all existing conversion methods are designed for CNNs, which adopt ReLU and Batch Normalization (BN) \cite{ioffe2015batch} for effective conversion. However, the standard ViT employs GELU and Layer Normalization (LN) \cite{ba2016layer}, which are incompatible with the computation paradigms of spike-based models and neuromorphic hardware, necessitating alternative solutions. On one hand, while GELU facilitates faster convergence and enhances model performance, its output conflicts with binary spikes in SNNs and violates the computational equivalence between ANNs and SNNs. Inspired by the initial Transformer \cite{vaswani2017attention}, which utilizes ReLU in MLP, we also adopt ReLU here to replace GELU for lossless conversion. On the other hand, despite the effectiveness of LNs in transformer-based models, its persistent normalization operations during inference remain incompatible with the event-driven computation paradigm of spike-based models. In contrast, BN offers a computational advantage, as it can be seamlessly absorbed into adjacent linear layers during inference, eliminating normalization overhead during inference. Also, it aligns well with fixed-sequence-length image data. Therefore, we substitute LNs with BNs, a strategy confirmed by recent studies \cite{zhou2022spikformer,yao2021leveraging}. Finally, we empirically implement BN after and ReLU before each linear layer. This design not only accelerates training convergence, but also confirms to requirements of spike-driven calculations after conversion.

\subsubsection{NReLU in Self-Attention}
\label{sec:nrelu}

MSA plays the most important role in transformers. It enables context-aware feature extraction by computing attention scores between tokens at different positions of a single sequence, thereby enhancing the model's ability to utilize task-relevant information while suppressing irrelevant components. For clarity, subsequent descriptions default to single-head attention (unless specified otherwise). Note that multi-head attention follows analogous computational procedures but employs multiple heads to capture comprehensive information, similar to multi-channel in CNNs.

Specifically, for the attention computation, the input $\boldsymbol{X}$ is first projected into query ($\boldsymbol{Q}$), key ($\boldsymbol{K}$) and value ($\boldsymbol{V}$) via distinct linear transformations, i.e., $\boldsymbol{Q}=\boldsymbol{XW_Q}$, $\boldsymbol{K}=\boldsymbol{XW_K}$ and $\boldsymbol{V}=\boldsymbol{XW_V}$. Subsequently, the attention scores are computed by measuring similarity between $\boldsymbol{Q}$ and $\boldsymbol{K}$, and then $\boldsymbol{V}$ is multiplied by this score to get the output,
\begin{equation}
\label{eq:sa}
    \mathrm{SA}(\boldsymbol{Q},\boldsymbol{K},\boldsymbol{V})=
    \boldsymbol{Attn}(\boldsymbol{Q},\boldsymbol{K}) \times \boldsymbol{V}=
    \mathrm{softmax}\left(\frac{\boldsymbol{Q} \times \boldsymbol{K}^T}{\sqrt{d_k}}\right) \times \boldsymbol{V}, \
    \mathrm{softmax}(x_i)=\frac{e^{x_i}}{\sum_{j}e^{x_j}}
\end{equation}
where $d_k$ is dimension of key matrix and $\times$ denotes matrix multiplication. $\mathrm{softmax}(x_i)$ is a normalization function, which is performed along the row dimension here.

From Eq.~(\ref{eq:sa}), we can see that conversion challenges of self-attention primarily lie in the softmax function, variable matrix multiplication (VMM) and constant scaling (e.g., $\frac{1}{\sqrt{d_k}}$). This section briefly introduces the solution for softmax, while remaining challenges are addressed in subsequent sections. The primary challenge for converting softmax stems from its inherent computational complexity, involving exponentiation and division, which fundamentally conflicts with the computation paradigm of SNNs and neuromorphic chips. To bridge this incompatibility, replacing softmax with simple yet effective alternatives is critical for fully spike-driven computation.

Compared to attention outputs normalized by softmax, we observe that outputs without softmax exhibit excessively large magnitudes, which critically contributes to training divergence (e.g., gradient explosion). To mitigate this, we use a scaling strategy. Inspired by the intrinsic normalization property of softmax, i.e., the expectation of each normalized output is $\frac{1}{N}$ (where $N$ is the sequence length), we adopt $N^{-1}$ as the universal scaling factor. With the importance of ReLU in ANN-to-SNN conversion, we further propose NReLU as a hardware-friendly alternative,
\begin{equation}
\label{eq:nrelu}
    \mathrm{NReLU}(x)=\mathrm{ReLU}(N^{-1} \cdot x)
\end{equation}
Experimental results validate that models with NReLU achieve competitive performance while eliminating hardware-unfriendly softmax operations. By substituting Eq.~(\ref{eq:nrelu}) into Eq.~(\ref{eq:sa}), and implementing modifications detailed in Sec.~\ref{sec:relu_bn}, we obtain the tailored self-attention (TSA):
\begin{equation}
\label{eq:tsa}
    \mathrm{TSA}(\boldsymbol{Q},\boldsymbol{K},\boldsymbol{V})=\mathrm{NReLU}\left(\frac{f(\boldsymbol{Q}) \times f(\boldsymbol{K}^T)}{\sqrt{d_k}}\right) \times f(\boldsymbol{V})
\end{equation}
where $f(\boldsymbol{X})=\mathrm{ReLU}(\mathrm{BN}(\boldsymbol{X}))$.

\subsection{Quantized Transformer}
\label{sec:quantformer}

For existing conversion methods, Quantization Clip-Floor-Shift (QCFS) technique bridges the computational gap between ANNs and SNNs through its quantization strategy, enabling SNNs to achieve high performance under ultra-low latency \cite{bu2023optimal}. The details of this method is shown in Appendix~\ref{sec:appendix_qcfs}. However, we find that models trained with QCFS exhibit non-convergence issues. We believe this is due to suboptimal parameter initialization, particularly the improper initialization of the clipping threshold $\lambda$. To address this limitation, we have chosen learned step size quantization (LSQ) technique \cite{esser2019learned} to quantize our models. The quantization functions of QCFS and LSQ are formulated as follows:
\begin{align}
    x_q&=\mathrm{QCFS}(x_f)=\lambda \cdot \mathrm{clip}\left(\frac{1}{L}\mathrm{round}\left(\frac{x_fL}{\lambda}\right),0,1\right) \\
    x_q&=\mathrm{LSQ}(x_f)=s\cdot\mathrm{round}\left(\mathrm{clip}\left(\frac{x_f}{s},0,L\right)\right)
\end{align}
where $x_f$ and $x_q$ represent the floating-point and fixed-point data, respectively. $\lambda$ and $s$ are learnable clipping thresholds, and $L$ is the quantization level. It can be shown that the two functions are computationally equivalent when $s=\lambda/L$. Unlike QCFS, which initializes $\lambda$ to a fixed value (typically 8 by default), LSQ adopts statistical initialization for $s$ and incorporates gradient scaling during training. These enhancements collectively enable faster convergence and enhanced training stability. To summarize, we systematically replace all ReLU activations (including in NReLU) in the tailored model with quantization functions, and perform LSQ to generate a quantized transformer.

\subsection{Spike-Driven Transformer}

\begin{figure}
    \centering
    \includegraphics[width=\textwidth]{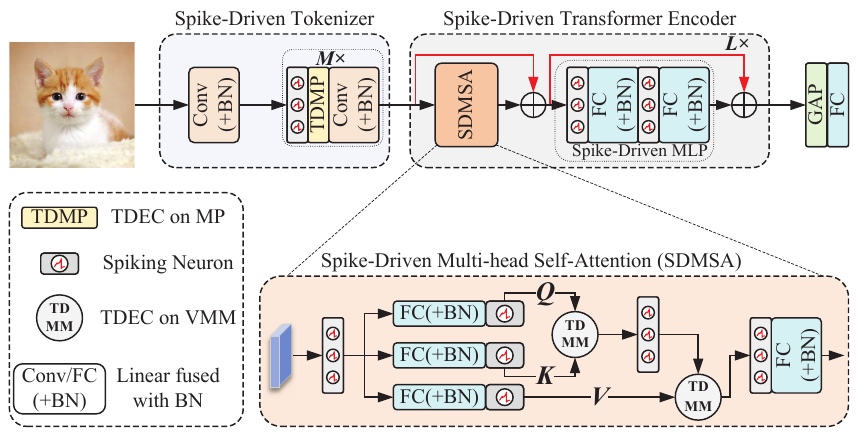}
    \caption{Architecture of Conversion Spike-Driven Transformer (CSDformer). Based on a quantized model, we first merge BNs into adjacent linear layers while absorbing constant scaling factors into thresholds, to avoid difficult calculations (e.g., division, square and square root). Second, temporal decomposition is applied to max-pooling and variable matrix multiplication for compatibility with temporal computations. Finally, we propose delayed Integrate-and-Fire neurons to replace all quantization functions, forming a fully spike-driven transformer. TDEC: temporal decomposition.}
    \label{fig:csdformer}
\end{figure}

At this stage, all necessary training procedures have been successfully finalized. This section outlines steps required to convert the quantized transformer into a spike-driven one before running. Before that, we adopt established technique in \cite{rueckauer2017conversion} to merge each BN, which is omitted in subsequent sections, into its adjacent linear layer. Fig.~\ref{fig:csdformer} shows the architecture of our CSDformer.

\subsubsection{Parameter Mapping}
\label{sec:param_map}

Regarding parameter mapping, we follow the strategy in \cite{bu2023optimal} and combine with the equivalence between LSQ and QCFS described in Sec.~\ref{sec:quantformer}, specifically mapping quantization level to time window (i.e., $T=L$), and clipping thresholds to firing thresholds (i.e., $\theta_l=s_lL$). Notably, we also address the challenge of scaling operations in self-attention here. Since these scaling factors remain temporally invariant, we strategically reconfigure firing thresholds to absorb these coefficients, i.e., $\theta^\prime_l=\theta_l/\delta$ (where $\delta=\frac{1}{\sqrt{d_k}}$ or $\frac{1}{N}$). This refinement eliminates constant multiplications while preserving the original charging and discharging dynamics of spiking neurons.

\subsubsection{Temporal Decomposition}
\label{sec:tdec}

So far, most calculations have been successfully converted to spike-driven operations, with the exception of max-pooling layers (MP), and VMMs in self-attention. To address these remaining challenges, we use Temporal DEComposition (TDEC) technique. Based on the theoretical proofs (see Appendix~\ref{sec:appendix_tdec}), we apply this technique to MP and VMM here.

\paragraph{TDEC on MP.}For simplicity, we focus on a single kernel of one MP layer in the network. Assume that the input and output are:
\begin{equation}
    \boldsymbol{x}=\mathrm{LSQ}(\mathrm{Conv}(\boldsymbol{x}_{pre}))=\frac{\boldsymbol{x}_T}{T}=\frac{\sum_1^T\boldsymbol{x}(t)}{T}, \quad y=\mathrm{MP}(\boldsymbol{x})=\frac{\mathrm{MP}(\boldsymbol{x}_T)}{T}=\frac{y_T}{T}
\end{equation}
where $\boldsymbol{x}_0=\boldsymbol{x}(0)=\boldsymbol{0}$ and $x_i(t)\in \{0, 1\}$. We formulate the TDEC operator for MP as $y(t)=\mathrm{MP}(\sum_0^t\boldsymbol{x}(t^\prime))-\mathrm{MP}(\sum_0^{t-1}\boldsymbol{x}(t^\prime))$, and subsequently establish $y_T=\mathrm{MP}(\boldsymbol{x}_T)=\sum_1^Ty(t)$. Fortunately, this decomposition ensures each temporal component is a binary spike, i.e., $y(t)\in \{0, 1\}$.

\paragraph{TDEC on VMM.}Combined with the description in Sec.~\ref{sec:quantformer}, we reformulate Eq.~(\ref{eq:tsa}) to quantize attention,
\begin{equation}
\label{eq:qsa}
    \mathrm{QSA}(\boldsymbol{Q,K,V})=f_q(\boldsymbol{Attn}) \times f_q(\boldsymbol{V})=f_q\left(\frac{f_q(\boldsymbol{Q}) \times f_q(\boldsymbol{K}^T)}{N\cdot\sqrt{d_k}}\right) \times f_q(\boldsymbol{V})
\end{equation}
where $f_q(\boldsymbol{X})=\mathrm{LSQ}(\boldsymbol{X})$. Sec.~\ref{sec:param_map} covered the conversion of constant scaling factors, we thus apply TDEC to address the conversion of $f_q(\boldsymbol{Q}) \times f_q(\boldsymbol{K}^T)$ and $f_q(\boldsymbol{Attn}) \times f_q(\boldsymbol{V})$ here. Considering the former as a case, we define variables as follows,
\begin{equation}
    \begin{split}
        &f_q(\boldsymbol{Q})=\frac{\boldsymbol{Q}_T}{T}=\frac{\sum_1^T\boldsymbol{Q}(t)}{T}, \quad 
        f_q(\boldsymbol{K})=\frac{\boldsymbol{K}_T}{T}=\frac{\sum_1^T\boldsymbol{K}(t)}{T} \\
        &f_q(\boldsymbol{Q}) \times f_q(\boldsymbol{K})=\frac{1}{T} \cdot \frac{\boldsymbol{Attn}_T}{T}=\frac{1}{T} \cdot \frac{\boldsymbol{Q}_T \times \boldsymbol{K}_T}{T}
    \end{split}
\end{equation}
where $\{Q_{ij}(t),K_{ij}(t)\} \in \{0,1\}$, and $\boldsymbol{K}^T$ is simplified to $\boldsymbol{K}$ to avoid the conflict between notations of time-window and transposition. In addition, $\frac{1}{T}$ in the above equation acts as a constant scaling factor, which can be absorbed into threshold as described in Sec.~\ref{sec:param_map}. Therefore, we also formulate the TDEC operator for VMM as:
\begin{equation}
    \boldsymbol{Attn}(t)=\boldsymbol{Q}_t \times \boldsymbol{K}(t)+\boldsymbol{Q}(t) \times \boldsymbol{K}_t-\boldsymbol{Q}(t) \times \boldsymbol{K}(t)
\end{equation}
where $\boldsymbol{Q}_t=\sum_1^t\boldsymbol{Q}(t^\prime)$ and $\boldsymbol{K}_t=\sum_1^t\boldsymbol{K}(t^\prime)$. Then, we get $\boldsymbol{Attn}_T=\boldsymbol{Q}_T \times \boldsymbol{K}_T=\sum_1^T\boldsymbol{Attn}(t)$ and each operator in $\boldsymbol{Attn}(t)$ is spike-driven. Similarly, $f_q(\boldsymbol{Attn}) \times f_q(\boldsymbol{V})$ can be decomposed using the same methodology.

\subsubsection{Spiking Neuron Model}
\label{sec:dif}
We have successfully implemented almost all necessary adjustments to convert the standard transformer, especially ViT, into its spike-driven counterpart. The final step is to substitute $\mathrm{LSQ}$ functions in the quantized model with spiking neurons. Although the quantization technology circumvents inherent clipping and quantization errors, it fails to address the issue of unevenness error. Fine-tuning after conversion is a typical solution \cite{wang2023toward,hu2023fast}. These methods improve performance but require substantial resources for retraining SNNs. Hao et al. developed training-free optimization strategies, however, they need strict conditions and complicated steps to succeed \cite{hao2023reducing,hao2023bridging}.

To maintain efficiency and simplicity, we propose the delayed Integrate-and-Fire (DIF) neuron model. This model ensures sufficient neural charging and discharging, and effectively reduces the error caused by the unevenness of input spikes. Algorithm~\ref{alg:delay_spike} shows the detailed workflow.
\begin{minipage}{0.8\linewidth}
    \begin{algorithm}[H]
        \renewcommand{\algorithmicrequire}{\textbf{Input:}}
        \renewcommand{\algorithmicensure}{\textbf{Output:}}
        \caption{Delayed Integrate-and-Fire Neuron}
        \label{alg:delay_spike}
        \begin{algorithmic}[1]
            \REQUIRE Input Spikes $s_j(t)$; Synaptic Weight $w_j$; Presynaptic Threshold $\theta_j$; Postsynaptic Threshold $\theta$; Initial Membrane Potential $v(0)=\theta/2$; Delayed Steps $\tau_d$; Time-window $T$.
            \ENSURE Output Spikes $s(t)$.
            \FOR{$t=1$ {\bfseries to} $\tau_d+T$}
            \IF{$t \leq T$}
            \STATE $v(t)=v(t-1)+\sum_jw_j\theta_js_j(t)$
            \ENDIF
            \IF{$t>\tau_d$}
            \STATE $s(t-\tau_d)=\Theta(v(t)-\theta)$
            \STATE $v(t)=v(t)-\theta s(t-\tau_d)$
            \ENDIF
            \ENDFOR
        \end{algorithmic}
    \end{algorithm}
\end{minipage}

\section{Experiments}
\label{sec:experiments}
We evaluate CSDformers on CIFAR-10/100 \cite{krizhevsky2009learning} and ImageNet-1K \cite{deng2009imagenet} datasets. Our experiments are implemented based on PyTorch \cite{paszke2019pytorch}, SpikingJelly \cite{spikingjelly} and Timm \cite{rw2019timm}. For the convenience of comparison, we follow the experimental setup in \cite{zhou2023spikingformer}. As for resources, we have trained models with eight A100 GPUs for ImageNet and a single one for CIFAR.

\paragraph{Performance on ImageNet.}Specifically, we adopt AdamW as the optimizer. The batch size is set to 288 during 310 training epochs, with a cosine-decay learning rate whose initial value is 0.0005. There are some convolutional modules with a total of four max-pooling layers in the tailored tokenizer, which splits the image into 196 patches of size $16\times16$ each.

Similar to Spikingformer, we conduct experiments on a variety of models with different embedding dimensions. The results are presented in Table~\ref{tb:imagenet}. Our CSDformers adhere to the spike-driven computation paradigm, effectively avoiding floating-point and fixed-point multiplications. Additionally, we compare the classification accuracy across different models, and our models demonstrate strong performance. On one hand, CSDformer shows consistent advantages across different configurations compared with directly trained models. CSDformer-8-384 achieves 75.55\% top-1 accuracy on ImageNet, significantly outperforming Spikformer-8-384 by 5.31\%, and surpassing Spikingformer-8-384 and Spike-driven Transformer-8-384 by 3.10\% and 3.27\%, respectively. CSDformer-8-512 reaches 76.18\% top-1 accuracy, leading Spikformer-8-512 by 2.80\%, with a 1.39\% and 1.61\% advantage over Spikingformer-8-512 and Spike-driven Transformer-8-512. The largest model CSDformer-8-768 maintains 76.36\% top-1 accuracy, matching the performance of Spike-driven Transformer-8-768 while exceeding Spikformer-8-768 and Spikingformer-8-768 by 1.55\% and 0.51\%. On the other hand, compared to conversion-based spiking transformers, CSDformer also achieves competitive performance under ultra-low timesteps. Notably, our models avoid difficult normalization operations, such as softmax and LNs, making them more hardware-friendly for neuromorphic devices.

\begin{table}[htbp!]
    \renewcommand{\arraystretch}{0.6}
    \setlength{\tabcolsep}{3pt}
    \caption{Results on ImageNet classification. Same as Spikingformer, our CSDformer-$L$-$D$ represents a CSDformer model with $L$ spiking transformer blocks and $D$ feature embedding dimensions. The default input resolution is $224\times224$. *The input crops are enlarged to $288\times288$ in inference.}
    \label{tb:imagenet}
    \begin{center}
    \begin{small}
    \begin{tabular}{cclcccc}
    \toprule
    \multirow{2}{*}{Category} & \multirow{2}{*}{Method} & \multirow{2}{*}{Architecture} & Spike   & Param & Time & Acc. \\
                              &                         &                               & -driven & (M)   & Step & (\%) \\
    \midrule
    \multirow{3}{*}{ANN}
    & ResNet \cite{yao2024spike}            & ResNet-104        & \XSolidBrush & 77.28 & 1 & 76.87 \\
    & Transformer \cite{zhou2022spikformer} & Transformer-8-512 & \XSolidBrush & 29.68 & 1 & 80.80 \\
    & Swin Transformer \cite{liu2021swin}   & Swin-B            & \XSolidBrush & 88.00 & 1 & 83.50 \\
    \midrule
    \multirow{13}{*}[-5ex]{\shortstack{SNN \\ (Direct \\ Training)}}
    & \multirow{2}{*}[-0.5ex]{SEW ResNet \cite{fang2021deep}} & SEW-ResNet-34  & \XSolidBrush & 21.79 & 4 & 67.04 \\
    &                                                 & SEW-ResNet-152 & \XSolidBrush & 60.19 & 4 & 69.26 \\
    \cmidrule{3-7}
    & \multirow{2}{*}[-0.5ex]{MS ResNet \cite{hu2024advancing}} & MS-ResNet-34   & \Checkmark & 21.79 & 6 & 69.42 \\
    &                                                   & MS-ResNet-104* & \Checkmark & 77.28 & 5 & 74.21 \\
    \cmidrule{3-7}
    &                                      & Spikformer-8-384 & \XSolidBrush & 16.81 & 4 & 70.24 \\
    & Spikformer \cite{zhou2022spikformer} & Spikformer-8-512 & \XSolidBrush & 29.68 & 4 & 73.38 \\
    &                                      & Spikformer-8-768 & \XSolidBrush & 66.34 & 4 & 74.81 \\
    \cmidrule{3-7}
    &                                            & Spikingformer-8-384 & \Checkmark & 16.81 & 4 & 72.45 \\
    & Spikingformer \cite{zhou2023spikingformer} & Spikingformer-8-512 & \Checkmark & 29.68 & 4 & 74.79 \\
    &                                            & Spikingformer-8-768 & \Checkmark & 66.34 & 4 & 75.85 \\
    \cmidrule{3-7}
    &                                              & Spiking Transformer-8-384  & \Checkmark & 16.81 & 4 & 72.28 \\
    & Spike-driven Transformer \cite{yao2024spike} & Spiking Transformer-8-512  & \Checkmark & 29.68 & 4 & 74.57 \\
    &                                              & Spiking Transformer-8-768  & \Checkmark & 66.34 & 4 & 76.32 \\
    \midrule
    \multirow{8}{*}[-5ex]{\shortstack{SNN \\ (Conversion)}}
    & QCFS \cite{bu2023optimal}           & VGG-16         & \Checkmark   & 138.36 & 256 & 74.22 \\
    & Fast-SNN \cite{hu2023fast}          & VGG-16         & \Checkmark   & 138.36 & 7   & 72.95 \\
    & Two-Stage \cite{wang2023toward}     & VGG-16         & \Checkmark   & 138.36 & 128 & 74.93 \\
    & MST \cite{wang2023masked}           & Swin-T (BN)    & \XSolidBrush & 28.50  & 256 & 78.37 \\
    & SpikeZIP-TF \cite{you2024spikezip}  & SViT-B-32Level & \XSolidBrush & 86.57  & 64  & 82.71 \\
    \cmidrule{3-7}
    &                                 & CSDformer-8-384 & \Checkmark & 16.81 & 4 & 75.55 \\
    & \textbf{This Work ($\tau_d=3$)} & CSDformer-8-512 & \Checkmark & 29.68 & 4 & 76.18 \\
    &                                 & CSDformer-8-768 & \Checkmark & 66.34 & 4 & 76.36 \\
    \bottomrule
    \end{tabular}
    \end{small}
    \end{center}
\end{table}

\paragraph{Performance on CIFAR.}For experiments on CIFAR, we follow the same setup for ImageNet with two exceptions: the batch size is set to 64 and max-pooling layers are included only in the last two convolutional modules of the tokenizer. The experimental results are summarized in Table~\ref{tb:cifar}. 

Overall, each variant of our models demonstrates a consistent performance advantage of 1\%-2\% over directly trained counterparts with identical architectures. Notably, the best-performing CSDformer-4-384-400E achieves 96.35\% classification accuracy on CIFAR-10, surpassing Spikformer-4-384-400E, Spikingformer-4-384-400E and Spike-driven Transformer by 0.84\%, 0.54\% and 0.75\%, respectively. On the more challenging CIFAR-100, the same architecture reaches 79.94\% accuracy, outperforming the aforementioned baselines by 1.73\%, 0.73\%, and 1.54\%. Moreover, conclusions from comparison of conversion-based models align with experiments on ImageNet.

\begin{table}
    \renewcommand{\arraystretch}{0.7}
    \setlength{\tabcolsep}{2.5pt}
    \caption{Results on CIFAR-10/100 classification. Same as Spikingformer, CSDformer-4-384-400E means CSDformer contains 4 spiking transformer blocks and 384 feature embedding dimensions, trained with 400 epochs. Also, other models of CSDformer are trained with 310 epochs by default. Note that Two-Stage \cite{wang2023toward} adopts VGG-16 for CIFAR-10 and ResNet-18 for CIFAR-100.}
    \label{tb:cifar}
    \begin{center}
    \small
    \begin{tabular}{cclcccccc}
    \toprule
    \multirow{2}{*}{Category} & \multirow{2}{*}{Method} & \multirow{2}{*}{Architecture} & Spike & Param & \multicolumn{2}{c}{CIFAR-10} & \multicolumn{2}{c}{CIFAR-100} \\
    \cmidrule(lr){6-7}\cmidrule(lr){8-9}
    & & & -driven & (M) & $T$ & Acc. & $T$ & Acc. \\
    
    \midrule
    \multirow{2}{*}{ANN}
    & ResNet \cite{zhou2022spikformer}      & ResNet-19         & \XSolidBrush & 12.63 & 1 & 94.97 & 1 & 75.35 \\
    & Transformer \cite{zhou2022spikformer} & Transformer-4-384 & \XSolidBrush & 9.32  & 1 & 96.73 & 1 & 81.02 \\
    
    \midrule
    \multirow{12}{*}[-5ex]{\shortstack{SNN \\ (Direct \\ Training)}}
    & STBP-tdBN \cite{zheng2021going}                           & ResNet-19 & \XSolidBrush & 12.63 & 6   & 93.16 & -   & -     \\
    & TET \cite{deng2022temporal}                               & ResNet-19 & \Checkmark   & 12.63 & 6   & 94.50 & 6   & 74.72 \\
    \cmidrule{3-9}
    & \multirow{4}{*}[-1.5ex]{Spikformer \cite{zhou2022spikformer}}
      & Spikformer-4-256      & \XSolidBrush & 4.15 & 4 & 93.94 & 4 & 75.96 \\
    & & Spikformer-2-384      & \XSolidBrush & 5.76 & 4 & 94.80 & 4 & 76.95 \\
    & & Spikformer-4-384      & \XSolidBrush & 9.32 & 4 & 95.19 & 4 & 77.86 \\
    & & Spikformer-4-384-400E & \XSolidBrush & 9.32 & 4 & 95.51 & 4 & 78.21 \\
    \cmidrule{3-9}
    & \multirow{4}{*}[-1.5ex]{Spikingformer \cite{zhou2023spikingformer}} 
      & Spikingformer-4-256      & \Checkmark & 4.15 & 4 & 94.77 & 4 & 77.43 \\
    & & Spikingformer-2-384      & \Checkmark & 5.76 & 4 & 95.22 & 4 & 78.34 \\
    & & Spikingformer-4-384      & \Checkmark & 9.32 & 4 & 95.61 & 4 & 79.09 \\
    & & Spikingformer-4-384-400E & \Checkmark & 9.32 & 4 & 95.81 & 4 & 79.21 \\
    \cmidrule{3-9}
    & Spike-driven & \multirow{2}{*}{Spiking Transformer} & \multirow{2}{*}{\Checkmark} & \multirow{2}{*}{9.32} & \multirow{2}{*}{4} & \multirow{2}{*}{95.60} & \multirow{2}{*}{4} & \multirow{2}{*}{78.40} \\
    & Transformer \cite{yao2024spike} & & & & & & & \\
    
    \midrule
    \multirow{9}{*}[-3ex]{\shortstack{SNN \\ (Conversion)}}
    & QCFS \cite{bu2023optimal}            & ResNet-18        & \Checkmark   & 11.69        & 64  & 96.06 & 64  & 79.54 \\
    & Two-Stage \cite{wang2023toward}      & VGG-16/ResNet-18 & \Checkmark   & -            & 4   & 94.06 & 4   & 75.98 \\
    & Fast-SNN \cite{hu2023fast}           & ResNet-18        & \Checkmark   & 11.69        & 7   & 95.57 & -   & -     \\
    & MST \cite{wang2023masked}            & Swin-T (BN)      & \XSolidBrush & 27.60        & 128 & 97.06 & 128 & 86.73 \\
    & SpikeZIP-TF \cite{you2024spikezip}   & SViT-S           & \XSolidBrush & 21.70        & 32  & 98.70 & 32  & 89.70 \\
    \cmidrule{3-9}
    & \multirow{4}{*}[-1.5ex]{\textbf{This Work ($\tau_d=3$)}}
      & CSDformer-4-256      & \Checkmark & 4.15 & 4 & 95.43 & 4 & 78.26 \\
    & & CSDformer-2-384      & \Checkmark & 5.76 & 4 & 95.52 & 4 & 78.45 \\
    & & CSDformer-4-384      & \Checkmark & 9.32 & 4 & 96.12 & 4 & 79.79 \\
    & & CSDformer-4-384-400E & \Checkmark & 9.32 & 4 & 96.35 & 4 & 79.94 \\
    \bottomrule
    \end{tabular}
    \end{center}
\end{table}

\begin{wraptable}{r}{8.5cm}
\setlength{\abovecaptionskip}{-0.1ex}
\setlength{\belowcaptionskip}{1ex}
\setlength{\tabcolsep}{5pt}
    \centering
    \caption{Training costs of CSDformer and Spikingformer on ImageNet. MACs and FLOPs for processing a single image are collected by calflops \cite{ye2023calflops}. Te: Training time per epoch; Tt: Total training time for 310 epochs.}
    \label{tb:cost}
    \begin{tabular}{*{5}{c}}
        \toprule
        \multirow{2}{*}{Model} & MACs & FLOPs & Te    & Tt     \\
                               & (G)  & (G)   & (min) & (days) \\
        \midrule
        Spikingformer-8-512 & 33.03 & 66.20  & 44 & 9.5  \\
        Spikingformer-8-768 & 74.05 & 148.32 & 77 & 16.6 \\
        CSDformer-8-512     & 8.26  & 16.55  & 22 & 4.7  \\
        CSDformer-8-768     & 18.51 & 37.08  & 23 & 5.0  \\
        \bottomrule
    \end{tabular}
\end{wraptable}

\paragraph{Training Cost.}We also evaluate the training time and resources, to illustrate efficiency benefits of CSDformer compared to directly trained models. As shown in Table~\ref{tb:cost}, we compared the costs of CSDformer and Spikingformer. Here, the training cost of CSDformer refers to that of the quantized model. The results show that CSDformer requires requires only a quarter of the computational resources compared to Spikingformer, and training speed is accelerated by 2-3$\times$. This is because the training of SNNs still uses floating-point precision on GPUs, and the temporal calculations introduce additional training overhead.

\section{Discussion}
\label{sec:discuss}

In addition to the above experiments, we conduct further analysis on neuron models in Appendix~\ref{sec:appendix_analyses}. Although these have demonstrated the superiority of the CSDformer, there are still several limitations that warrant improvements: \textbf{\emph{Neuromorphic Vision Tasks.}} When CSDformer is applied to neuromorphic datasets for further energy efficiency, its performance becomes unsatisfactory. We attribute this limitation to the quantized model's inability to capture spatial-temporal features, and future research should explore more effective pre-processing and training strategies. \textbf{\emph{Inference Computational Complexity.}} Our model employs TDEC technique, whereas directly trained models inherently leverage spikes for inference and have significantly lower complexity, inspiring us to explore streamlined conversion methods and spike-driven architectures. \textbf{\emph{Scalability and Applications.}} Current spiking transformers, including our CSDformer, are primarily limited to small architectures and classification tasks. Scaling to larger networks and complex applications intensifies GPU resource demands, necessitating training-free conversion methods, especially for privacy or massive datasets.

\section{Conclusion}
\label{sec:conclusion}
In this paper, we propose a novel conversion framework for fully spike-driven transformers, named CSDformer, to avoid training SNNs and hardware-unfriendly calculations. We redesign core modules of a standard transformer and train its quantized version, which is then converted into a spike-driven model. This method not only replaces inefficient computations in the original architecture into operations compatible with neuromorphic hardware, but also significantly reduces training costs. We evaluate CSDformer on ImageNet, CIFAR-10 and CIFAR-100 datasets. Experimental results show the consistent superiority of CSDformer across various aspects (e.g., classification accuracy, training costs and computational resources), and our model outperforms directly trained spike-based transformers by a substantial margin. To the best of our knowledge, CSDformer is the first fully spike-driven transformer generated through conversion method, and achieves excellent performance under ultra-low latency, while drastically reducing computational complexity and training costs. We hope that our investigation will pave the way for further research on spike-driven transformers, and inspire the design of next generation neuromorphic chips.


\newpage
{
\small
\bibliographystyle{csdformer}
\bibliography{csdformer}
}


\newpage
\appendix

\section{Quantization Clip-Floor-Shift Method}
\label{sec:appendix_qcfs}

\paragraph{Activation for ANNs.}In each layer, the output is determined by performing linear transforms and applying a non-linear activation. Usually, the activation function for ANNs is selected as ReLU for conversion, which is similar to the behavior of Integrate-and-Fire (IF) neurons in SNNs:
\begin{equation}
    \boldsymbol{a}_l=\mathrm{ReLU}(\boldsymbol{W}_l\boldsymbol{a}_{l-1})=\mathrm{max}(\boldsymbol{W}_l\boldsymbol{a}_{l-1},\boldsymbol{0}), \quad l=1,2,...,L
\end{equation}
where $\boldsymbol{a}^l$ is the output, and $\boldsymbol{W}^l$ is the weight matrix connected to neurons in $l$-th layer. For simplicity, the bias term is omitted.

\paragraph{Activation for SNNs.}As mentioned before, the IF neuron model is used in existing conversion methods. Each neuron undergoes charging and discharging process,
\begin{equation}
\label{eq:if_vector}
    \begin{aligned}
        & \boldsymbol{u}_{l}(t)=\boldsymbol{v}_{l}(t-1)+\boldsymbol{I}_{l-1}(t) \\
        & \boldsymbol{s}_{l}(t)=\Theta(\boldsymbol{u}_{l}(t)-\theta_{l}) \\
        & \boldsymbol{v}_{l}(t)=\boldsymbol{u}_{l}(t)-\boldsymbol{x}_{l}(t)
    \end{aligned}
\end{equation}
where $\boldsymbol{I}_{l-1}(t)=\boldsymbol{W}_l\boldsymbol{x}_{l-1}(t)$ is the input of neurons in $l$th layer. $\boldsymbol{x}_{l}(t)=\boldsymbol{s}_{l}(t)\theta_l$ is the unweighted postsynaptic potential, which is also the input of next layer. 

\paragraph{ANN-to-SNN Conversion.}By iterating Eq.~(\ref{eq:if_vector}) and averaging it over time window ($T$), we have:
\begin{equation}
\label{if_iter}
    \frac{\boldsymbol{v}_l(T)-\boldsymbol{v}_l(0)}{T}=\boldsymbol{W}_l
    \frac{\sum_1^T\boldsymbol{x}_{l-1}(t)}{T}-
    \frac{\sum_1^T\boldsymbol{x}_{l}(t)}{T}
\end{equation}
If we define $\boldsymbol{\phi}_l=\frac{\sum_1^T\boldsymbol{x}_{l}(t)}{T}$ as the average potential, Eq.~(\ref{if_iter}) can be rewritten as:
\begin{equation}
    \boldsymbol{\phi}_l=\boldsymbol{W}_l\boldsymbol{\phi}_{l-1}-\frac{\boldsymbol{v}_l(T)-\boldsymbol{v}_l(0)}{T}
\end{equation}

Due to the binary spikes, the value of $\boldsymbol{\phi}_{l}$ falls within \{$0,\frac{\theta_l}{T},...,\frac{(T-1)\theta_l}{T},\theta_l$\}, while $\boldsymbol{a}_{l}$ is unbounded. Therefore, to achieve more accurate equivalent mapping, authors replace ReLU with quantization clip-floor (QCF) activation function during training to limit the output,
\begin{equation}
    \boldsymbol{a}_l=\lambda_l\mathrm{clip}\left(
    \frac{1}{L}\left\lfloor \frac{\boldsymbol{W}_l\boldsymbol{a}_{l-1}L}{\lambda_l} \right\rfloor,0,1
    \right)
\end{equation}
where $L$ denotes quantization steps, the learnable $\lambda_l$ decides the maximum value of $\boldsymbol{a}_l$. It shows that the theoretical error between the converted SNN and the quantized ANN is zero. They further proposed QCFS function,
\begin{equation}
\label{eq:qcfs}
    \boldsymbol{a}_l=\lambda_l\mathrm{clip}\left(
    \frac{1}{L}\left\lfloor \frac{\boldsymbol{W}_l\boldsymbol{a}_{l-1}L}{\lambda_l}+\frac{1}{2} \right\rfloor,0,1
    \right)
\end{equation}
It shows that this method not only avoids to retrain ANN when SNN running with different time window, but also ensures the expected error is zero and the expected square error is minimal.

\section{Proofs of Temporal Decomposition}
\label{sec:appendix_tdec}

\subsection{Temporal Decomposition on Non-linear Function.}
Assume that for a given non-linear function $f$, the following conditions hold,
\begin{equation}
    x_t=\sum_1^tx(t^\prime), \quad y_t=f(x_t) \ \mathrm{and} \ y_0=f(x_0)=0
\end{equation}

Obviously, $y_t=f(\sum_1^tx(t^\prime))\neq\sum_1^tf(x(t^\prime))$. To align with the temporal computing paradigm in SNNs, $y_t$ should be decomposed into its temporal components. We thus apply Temporal Decomposition to $f(x_t)$,
\begin{equation}
    \begin{split}
        y_t&=f(x_t)\\
        &=f(x_t)-f(x_{t-1})+\dots+f(x_1)-f(x_0)\\
        &=\sum\nolimits_1^t(f(x_{t^\prime})-f(x_{t^\prime-1}))
    \end{split}
\end{equation}
By setting $y(t^\prime)=f(x_{t^\prime})-f(x_{t^\prime-1})=f(\sum_0^{t^\prime}x(t^{\prime\prime}))-f(\sum_0^{t^\prime-1}x(t^{\prime\prime}))$, we derive that $y_t=f(x_t)=\sum_1^ty(t^\prime)$.

\subsection{Temporal Decomposition on Matrix Multiplication.}
Assume that the following equations are valid,
\begin{equation}
    \boldsymbol{A}_t=\sum_1^t\boldsymbol{A}(t^\prime), \quad \boldsymbol{B}_t=\sum_1^t\boldsymbol{B}(t^\prime), \quad \boldsymbol{C}_t=\boldsymbol{A}_t \times \boldsymbol{B}_t
\end{equation}
 Also, we should decompose $\boldsymbol{C}_t$, and thus apply Temporal Decomposition to $\boldsymbol{C}_t$,
\begin{equation}
    \begin{split}
        \boldsymbol{C}_t&=\boldsymbol{A}_t \times \boldsymbol{B}_t=\sum_1^t\boldsymbol{A}(t^\prime) \times \sum_1^t\boldsymbol{B}(t^\prime)\\
        &=\sum_1^t(\boldsymbol{A}_{t^\prime} \times \boldsymbol{B}(t^\prime)+\boldsymbol{A}(t^\prime) \times \boldsymbol{B}_{t^\prime}-\boldsymbol{A}(t^\prime) \times \boldsymbol{B}(t^\prime))
    \end{split}
\end{equation}
By setting $\boldsymbol{C}({t^\prime})=\boldsymbol{A}_{t^\prime} \times \boldsymbol{B}(t^\prime)+\boldsymbol{A}(t^\prime) \times \boldsymbol{B}_{t^\prime}-\boldsymbol{A}(t^\prime) \times \boldsymbol{B}(t^\prime)$, we derive that $\boldsymbol{C}_t=\boldsymbol{A}_t \times \boldsymbol{B}_t=\sum_1^t\boldsymbol{C}({t^\prime})$. Notably, when $\boldsymbol{A}(t^\prime)$ and $\boldsymbol{B}(t^\prime)$ are both binary spikes, each item in $\boldsymbol{C}(t^\prime)$ also aligns with the spike-driven paradigm.

\section{Further Analysis of Neuron Model}
\label{sec:appendix_analyses}

In this paper, we proposed DIF neurons to replace the commonly used IF neurons, to address the issue of unevenness error (see Sec.~\ref{sec:dif}). To validate the effectiveness of DIF model, we evaluate the impact of varying delayed time-steps on accuracy. As shown in Table~\ref{tb:dif}, the results reveal that models employing standard IF neurons exhibit suboptimal performance, or even fail to function effectively. Meanwhile, the performance of models with DIF neurons progressively improves as the number of delayed steps increases. When the steps reach a certain value, neurons achieve sufficient charging and discharging, and the performance no longer improves.

\begin{table}[htbp!]
    \centering
    \caption{Results with different delayed steps. We test CSDformer-4-384 on CIFAR-100 and CSDformer-8-768 on ImageNet. The DIF neuron with zero delayed steps is equivalent to IF neuron.}
    \label{tb:dif}
    \begin{tabular}{*{4}{c}}
        \toprule
        Model & Neuron & Delay ($\tau_d$) & Acc. (\%) \\
        \midrule
        \multirow{4}{*}{\shortstack{CSDformer-4-384\\($T=4$)}}
        & IF  & 0       & 67.76          \\
        & DIF & 1       & 75.65          \\
        & DIF & 2       & 79.12          \\
        & DIF & $\ge$ 3 & \textbf{79.79} \\
        
        \midrule
        \multirow{4}{*}{\shortstack{CSDformer-8-768\\($T=4$)}}
        & IF  & 0       & 2.21           \\
        & DIF & 1       & 69.61          \\
        & DIF & 2       & 72.37          \\
        & DIF & $\ge$ 3 & \textbf{76.36} \\
        \bottomrule
    \end{tabular}
\end{table}

\end{document}